# A Cost-Effective Approach to Smooth A* Path Planning for Autonomous Vehicles

Lukas Schichler*, Karin Festl*, Selim Solmaz* and Daniel Watzenig* †
*Virtual Vehicle Research GmbH, Graz, Austria
Email: lukas.schichler@v2c2.at
†Institute of Computer Graphics and Vision, TU Graz

*Abstract*—Path planning for wheeled mobile robots is a critical component in the field of automation and intelligent transportation systems. Car-like vehicles, which have non-holonomic constraints on their movement capability impose additional requirements on the planned paths. Traditional path planning algorithms, such as A*, are widely used due to their simplicity and effectiveness in finding optimal paths in complex environments. However, these algorithms often do not consider vehicle dynamics, resulting in paths that are infeasible or impractical for actual driving. Specifically, a path that minimizes the number of grid cells may still be too curvy or sharp for a car-like vehicle to navigate smoothly.

This paper addresses the need for a path planning solution that not only finds a feasible path but also ensures that the path is smooth and drivable. By adapting the A* algorithm for a curvature constraint and incorporating a cost function that considers the smoothness of possible paths, we aim to bridge the gap between grid based path planning and smooth paths that are drivable by car-like vehicles. The proposed method leverages motion primitives, pre-computed using a ribbon based path planner that produces smooth paths of minimum curvature. The motion primitives guide the A* algorithm in finding paths that are optimal in terms of both length and curvature. With the proposed modification on the A* algorithm, the planned paths can be constraint to have a minimum turning radius much larger than the grid size. We demonstrate the effectiveness of the proposed path planning algorithm in different unstructured environments. In a two-stage planning approach, first the modified A* algorithm finds a grid-based path and the ribbon based path planner creates a smooth path within the area of grid cells. The resulting paths are smooth with small curvatures independent of the orientation of the grid axes and even in presence of sharp obstacles.

## I. Introduction

Autonomous vehicle applications in unstructured environments include rescue, construction site, parking lots etc. and in many of these applications, the vehicle perceives its surrounding and creates a map of the environment using SLAM [1, 2]. To reach its target, dedicated planning algorithms are required. A review on path planning approaches can be found in [3, 4, 5]. There are numerous approaches simultaneously dealing with the non-holonomic constraints of the vehicle and solving the path search problem in the unstructured, possibly not fully known, environment. These solutions range from classic geometric approaches based on Reeds and Shepp [6] to game-theoretic and AI based approaches [7]. A popular solution is the RRT path planner [3, 2, 8]. All these approaches face the challenge of high computational effort with complex environment structures. An alternative to these are grid based path planning algorithms. At the cost of disregarding the vehicles movement restrictions, grid based search algorithms can provide fast solutions even in complex environments. One of the most famous is the A* algorithm.

Grid based algorithms find a series of grid cells that lead to the target, optimizing a given cost. Simply connecting the solution grid cells leads to a path with sharp turns. Thus, further steps are required to obtain a path that is drivable by car-like vehicles. One way to reduce sharp turns is using a hexagonal grid instead of an orthogonal grid (reducing the turn angle from $90 \deg$ to $60 \deg$). The resulting path can then be smoothed using different geometric algorithms as is described in [9]. Interpolation methods can create paths of arbitrary smoothness, however collision checks need to be implemented in addition to the interpolation method which leads to high computational costs. Moreover, while the degree of smoothness is defined directly, the maximum curvature needs to be evaluated and adapted with numeric methods. There are also path smoothing methods using special curves such as the Dubins path (e.g. [10]), which can handle constraints such as obstacles and maximum curvature, but cannot optimize e.g. for the curvature. Therefore, in this work, we use a heuristical optimization based path planner to create a smooth path in the selected grid cells that satisfies constraints on the maximum curvature and the path boundaries. Due to this curvature constraint, the cell size can be decreased to reflect obstacles in more detail.

Besides constraining the curvature, it is also beneficial to add a cost function that penalizes curvy paths. There are already solutions to regard a turning cost, as for example in [11]. In these approaches, the assumption is that a change

This paper is part of the AI4CSM project that has received funding from the ECSEL Joint Undertaking (JU) under grant agreement No 101007326. The JU receives support from the European Union's Horizon 2020 research and innovation programme and Germany, Austria, Belgium, Czech Republic, Italy, Netherlands, Lithuania, Latvia, Norway". In Austria the project was also funded by the program "IKT der Zukunft" of the Austrian Federal Ministry for Climate Action (BMK). The publication was written at Virtual Vehicle Research GmbH in Graz and partially funded within the COMET K2 Competence Centers for Excellent Technologies from the Austrian Federal Ministry for Climate Action (BMK), the Austrian Federal Ministry for Labour and Economy (BMAW), the Province of Styria (Dept. 12) and the Styrian Business Promotion Agency (SFG). The Austrian Research Promotion Agency (FFG) has been authorised for the programme management.

of direction from one pair of connected cells to the next, causes a specific curvature in the reference path. However, the actual smoothed path can have any shape that fits into the area of selected cells. For example, moving horizontally in a hexagonal grid should not induce a curvy path despite the changes of directions in the grid path. An example to this is shown in Fig. 1. In our approach, we couple the A* grid search algorithm with a smooth ribbon based path planning algorithm [12]. We define solution graph primitives consisting of $n$ interconnected cells in all admissible configurations. For these primitives, smooth path solutions with constrained curvature are pre-computed using a heuristic ribbon planner optimizing for minimum curvature. The A* algorithm is modified to reach only for cells that match these primitives. For this, the cell costs are computed and stored in a new graph structure implementing the predefined primitives.

The presented approach is applied on a hexagonal grid based A* algorithm with a primitive length of $n = 5$. The primitives are designed such that only paths which allow a curve radius that is large compared to the grid cell size. This allows for a fine resolution of obstacles, even though the curve radius of the vehicle is large. The primitives are shown in Fig. 2. They are defined as a sequence of changes of directions. The shown paths can be rotated and mirrored to reproduce all possible directions. With this approach, the problem of tight cornering around edges is solved. Moreover the problem of higher costs in directions that are not aligned with the coordinate axis is solved.

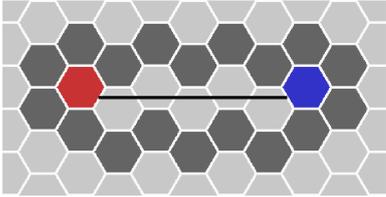

Fig. 1: Moving horizontally in a hexagonal grid produces a zig-zag in the grid cell formation. Nevertheless, a straight path can be planned within these grid cells.

## II. GRID BASED PATH PLANNING WITH LIMITED CURVATURE

The algorithm presented in this work is based on a hexagonal grid, which means that each cell is connected to 6 adjacent cells. Without restriction, a grid based path can include turns of $180 \deg$ (the path leads from cell A to cell B and back to cell A again), turns of $120 \deg$ and $60 \deg$, and straight lines. A vehicle with limited turning radius (e.g. any car-like vehicle with limited steering angle) may be unable to follow these paths. For this reason, in this section, we introduce restrictions for path candidates.

A grid based path is considered feasible, if a continuous path satisfying the constraints on the curvature can be designed within the area of the grid based path, leading from the start cell to the target cell. The feasibility of a grid based path depends on the formation of the path cells and the ratio between inner cell radius $r_c$ and minimum turning radius $r_{\min}$. For a better understanding, consider 3 cases:

- c1 $r_{\min}/r_c \leq 1$: the vehicle can fully turn within one cell. Any formation is feasible.
- c2 $1 < r_{\min}/r_c \leq \sqrt{7}$: a full turn of the vehicle fits into 6 cells forming a circle. The grid based path is feasible if it does only contain turns of $60 \deg$ and straight lines.
- c3 $\sqrt{7} < r_{\min}/r_c \leq 3.329$: a full turn of the vehicle fits into 12 cells forming a circle. The grid based path is feasible if it includes only turns of $60 \deg$ followed by a straight line or a turn of $60 \deg$ in the other direction. Feasible paths are shown in Fig. 2.

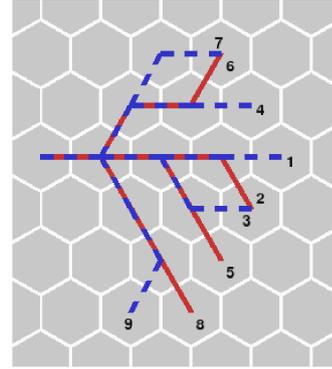

Fig. 2: Path primitives with a length of $n = 5$ cells representing all admissible path formations regarding the curvature constraint. Each primitive starts on the left and ends at their respective number.

A large cell size (case c1) has the benefit of not requiring any restrictions on the cell formation as well as leading to low computational effort due to the low number of cells. However, the environment (especially small obstacles and narrow pathways) cannot be represented by the grid accurately. For smaller cell size, restrictions on the formation of 3 cells (case c2) and 5 cells (case c3) emerge. The cell size can be further reduced to arbitrary values leading to restrictions on formations of more cells. These restrictions lead to additional computational effort, as will be demonstrated for the A* path planner in the next section. We will present the modifications for case c3 and consider $r_{\min}/r_c = 3.329$ for the rest of this work.

## III. A* WITH LIMITED CURVATURE

The path planning algorithm presented in this work is based on the A* algorithm on a hexagonal grid with modifications to consider a limited curvature. We first introduce the concept of the modification and afterwards describe the algorithm in detail.

### A. The concept

The standard A* algorithm builds a tree of paths from the start cell in directions of optimal cost. In each iteration, the paths of lowest cost are extended by adding new admissible cells (so called open cells) to their end cell (leaf cell). Without restrictions on the curvature, the list of open cells includes all cells that are not part of the tree, are adjacent to a leaf cell and are not occupied by an obstacle (free cell).

With restrictions on the curvature, the admissibility of a cell depends on the shape of the path leading to this cell. Therefore, in the tree of paths built by the modified A* algorithm, each cell can have multiple nodes. In other words, for each open cell, there is a list of admissible paths leading to this cell and a cost value for each path. To check if an adjacent cell is admissible, the algorithm propagates these paths back to check if it is part of the primitives shown in Fig. 2.

An exemplary state of the modified A* algorithm is shown in Fig. 3. This figure and all oncoming figures are created using the visualization library pygame, with the color scheme defined in Tab. I. The arrows between cells indicate the path tree structure, pointing from parent to child. Lighter arrows indicate that multiple paths include the corresponding two cells. The hexagonal grid coordinate system used in our examples is axial, meaning that the first axis points from left to right and the second axis from top left to bottom right. This coordinate system approach also used in [13] and is inspired by the cube coordinate system without the extra axis.

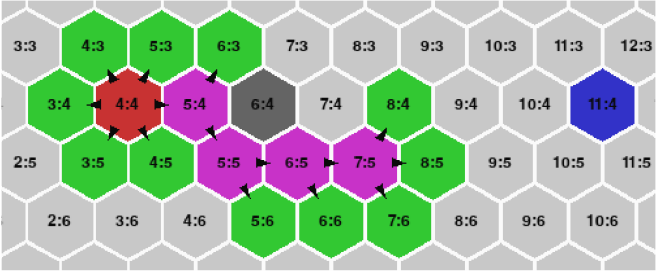

Fig. 3: Exemplary state of the modified A* algorithm. Cells are added to the open-list (green) only when the curvature constraints are satisfied.

TABLE I: Color scheme for the visualization

| Color | Cell type |
|---|---|
| light grey | free cell |
| dark grey | occupied cell |
| red | start cell |
| blue | target cell |
| green | open cell |
| pink | closed cell |

In Fig. 3, the cell 7:4 is neither reachable from 6:5 nor 7:5, as the curvature constraint is not satisfied. After further progression, this cell may be admissible through 5:6 → 6:5 → 7:4, making 6:5 open again.

### B. The algorithm in detail

In the initial step, the start cell is the current cell. In each iteration, the following steps are executed:

1) Add each cell adjacent to the start cell to the open-list. An open-list entry is composed of the cell index, a list of paths leading to that cell and the cost $c$ for each of these paths. In the initial step, the list of paths is only one path composed of only the start cell. Computation of the cost $c$ is described in the next section. As long as the path including the new cell is shorter than the primitives, the curvature cost $c_\kappa$ is 0.
2) From all paths of all cells in the open-list, select the cell with the cheapest path. This cheapest path is from now on considered a closed-path. If a cell has no more open paths, it is considered a closed cell.
3) Append the selected cell to the cheapest path leading to that cell.
4) Based on the shape of the cheapest path, determine adjacent cells which comply with the curvature constraint.
5) Add all admissible cells adjacent to the selected cell to the open-list. For each new open cell, the cheapest path extended in step 3 is added to the list of paths. If the cell is already in the open-list, only the path is added. The cost $c$ is computed as described in the next section. For the cost penalizing curvature $c_\kappa$, the new open cell and the last 4 entries from the path are considered.
6) Unless one of the new cells is the target cell, continue with step 3.
7) The target cell is in the open-list with corresponding paths leading from the start cell to the target cell. Select the path with the lowest cost.

Further, to reduce the number of iterations, the following modifications are implemented:

- If multiple similar paths (the past 3 cells are identical) lead to the cheapest cell selected in step 2, append all of them to that cell and calculate their costs respectively.
- When a cell has no admissible adjacent cells (e.g. when it is surrounded by obstacles) it is marked as a dead cell, preventing it from becoming an open cell again. Yellow arrows indicate a found dead end.

## IV. COST FUNCTION

To decide which open cell is added to the tree of admissible paths, thus transformed to a closed cell, a cost for each open cell is computed. This cost $c$ is composed as follows [3]:

$$c = c_c + c_g \quad (1)$$

Where $c_c$ is the cost-to-come and $c_g$ is the cost-to-go. The cost-to-come is an estimate of the lowest cost to get from the start cell to the open cell. In our approach it is approximated by the weighted sum of the number of cells of the cheapest path $n_c^*$ from the start cell and a cost $c_\kappa$ corresponding to the curvature of the path primitive leading to the open cell.

$$c_c = w_n \cdot n_c^* + w_\kappa \dot c_\kappa \quad (2)$$

The number of cells $n_c^*$ can be computed incrementally. For the cost $c_\kappa$, for each path primitive, a fixed cost value is pre-computed as described later. The effective path primitive is evaluated during the check for admissibility in step 4. The cost-to-go is approximated by the euclidean distance to the target cell. By weighing the different parts of the cost function, the solution of the cheapest path will change and also the trade-off between exploration and exploitation can be tuned.

For each primitive shown in Fig. 2, the cost penalizing path curvature is precomputed. For this, we implemented 3 variants described below. The resulting cost values are shown in Tab. II.

### A. Ribbon

A ribbon model based path planning algorithm [12] is used to plan a smooth path within the cell area of each primitive. The ribbon is constructed of the edges of all cells in the primitive. The planning algorithm iteratively computes circular path sections forming a continuous path within the ribbon satisfying constraints on the curvature. The implemented heuristic aims on minimizing the curvature of the planned path. The cost value $c_\kappa$ for each primitive is the median of the planned path curvature.

### B. Adapted Ribbon

For the ribbon model based path planner, the start position and orientation is fixed, while the goal position and orientation is free. This leads to different costs when a primitive is paced in one direction or the other. To counteract this effect, the costs of primitives representing the more expensive direction are set to their respective lower cost. This is the case for the primitive 8 (as labeled in Fig. 2, which is set to the value of primitive 2, as well as primitive 4 which is set to the value primitive 3. Additionally, with a proper start orientation, the cost of primitive 6 should match the cost of a straight line (primitive 1).

### C. Curvature Penalty

To tune the path planning behavior to specifically optimize for a small curvature and a small number of changes of direction, the cost distributions as set manually. The cost values of primitives 5, 7 and 9 represent the largest curvature and where therefore set to maximum.

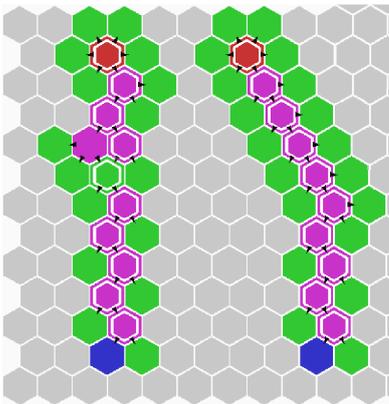

Fig. 4: Algorithm performing on an open map, planning straight lines in different angles.

TABLE II: Curvature costs $c_\kappa$ for each path primitive

| ID | Ribbon | Adapted Ribbon | Curvature Penalty |
|----|--------|----------------|-------------------|
| 1  | 0.0    | 0.0            | 0.0               |
| 2  | 0.087  | 0.087          | 0.1               |
| 3  | 0.119  | 0.119          | 0.2               |
| 4  | 0.195  | 0.119          | 0.2               |
| 5  | 0.429  | 0.429          | 1.0               |
| 6  | 0.109  | 0.0            | 0.0               |
| 7  | 0.429  | 0.429          | 1.0               |
| 8  | 0.507  | 0.087          | 0.1               |
| 9  | 0.915  | 0.915          | 1.0               |

## V. EVALUATION

With the presented modification of the A* algorithm, it is possible to find paths in a grid, satisfying constraints on the curvature (with a turning radius greater than the cell size) and minimizing the curvature of the path. By evaluating different scenarios, we want to show the following properties of the presented algorithm:

E1 Even though the target cell is positioned in a direction that is not aligned with any of the grid axes, the planned path is shaped such that a (close to) straight path can be planned within the selected grid cells.

E2 In presence of obstacles with sharp edges and complicated structures the planned path bypasses the occupied cells satisfying the maximum curvature constraint.

E3 By tuning the parameters for the curvature cost, different behaviors can be achieved. The path can be tuned for a short path, small curvature or a small number of changes of direction.

For the evaluation, the cost weights were set to $w_n = 1.0, w_\kappa = 5.0$ and the curvature cost distribution *Curvature Penalty* of Tab. II was used.

### E1. Open map

Straight paths in directions that are not aligned with the grid axis result in curvy grid-based paths. An appropriate cost function is low for path primitives in formations where a straight smooth path can be planned within the cells. These are formations involving 2 consecutive changes of directions as in primitives 3 and 6.

In Fig. 4 the algorithm planning different angled straight lines in an open map is shown. With the costs function IV-C the algorithm prefers straight line along the axis first before taking the shortest path downwards. The reason is that, to switch from primitive 1 to 6, the path first includes the primitives 2 and 3, which both have higher costs.

### E2. Curvature constraint

In Fig. 5 and Fig. 6, two characteristics of the modified A* algorithm are visualized.

First, even in limited space, the curvature constraint is satisfied. As demonstrated in Fig. 5, the algorithm creates a turning maneuver to escape narrow spaces. Right after the escape, the shortest path straight down would contradict the curvature constraint, thus a curve with a larger radius is planned.

Second, due to the fact that cells can be part of multiple paths, the number of iterations to explore paths directed to the target drastically increases when the path is blocked, as in Fig. 6. Cells that are part of multiple paths are indicated with white and light-gray arrows. To counter act this problem, dead cells as described in Sec.III-B are introduced. They are indicated with yellow arrows. As soon as a cell is detected as dead-cell, it can no longer become an open cell.

*E3. Parameter tuning*

In general, tuning parameters for the A* algorithm can significantly impact its performance and efficiency. As the focus of this paper is on the curvature constrains and path smoothness, the parameters tuned are the curvature cost distributions. In Tab. II the cost distribution of three different cost functions are shown. In Fig. 7, the results of these three cost distributions are shown. In the resulting grid-based path, a smooth path is created using the ribbon based planner described in Sec. IV-A is used.

With the cost distribution computed by the ribbon based planner (see Fig. 7a) results in a short path with sharp turns exhausting the curvature constraint. With adaptions to this cost distribution, the curvature decreases (see Fig. 7b). With the cost distribution manually set to minimize the curvature (see Fig. 7c, the resulting path becomes even longer, creating a path of small curvature that is close to constant along the whole path.

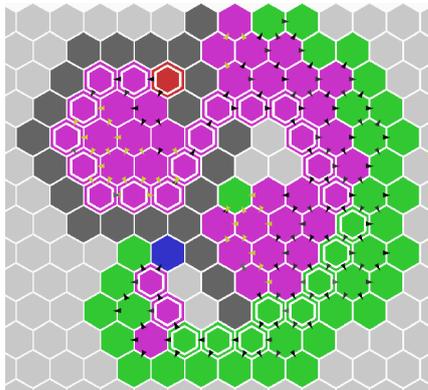

Fig. 5: To escape a narrow space while satisfying the curvature constraint, the algorithm creates a turning maneuver.

## VI. CONCLUSION

The A* algorithm is a long approved and well established method for path planning for mobile robots in unstructured environments. To create a path feasible for car-like vehicles or high demands on dynamic performance, a 2-stage planning approach can be a good solution. Based on the grid based path planned by the A* algorithm, a smooth, drivable path is computed by a second planning algorithm. With this approach, several challenges arise:

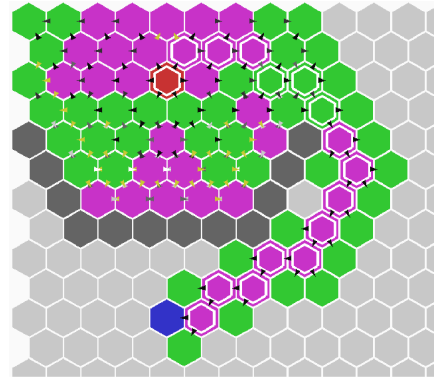

Fig. 6: When the path directed to the target cell is blocked, the number of iterations increases as multiple paths are planned in each cell.

- The unstructured environment has to be represented by a grid. If the grid is too coarse, the boundaries of obstacles may need to be highly overestimated.
- The grid based path, consisting of adjacent cells, may be in a formation such that it is not possible to plan a drivable path within the selected cells. Especially with a fine grid, constraints on the curvature of the smooth path may not be satisfied.
- A grid can only render straight paths in directions aligned with the grid axes. For example in an orthogonal grid, multiple paths of similar length (number of cells) lead to a target in diagonal direction. It is therefore difficult to find a good measure for path length and curviness.

To address these challenges, we proposed a modification of the A* algorithm consisting of two parts: first, the implementation of a hard constraint on the curvature and second, a cost function that represents the curviness of an optimal smooth path within the grid based path.

With the hard constraints on the curvature of the path, drivability of the resulting smooth path is guaranteed, even though the grid cells are much smaller than the minimum curve radius. These constraints are implemented by defining a set of primitives that are considered admissible. For the same set of primitives, we planned a smooth path within the are of the cells. Evaluating the curvature of these smooth paths provides an appropriate measure for the curviness of the resulting path, making it a good solution for a cost function that penalizes path enabling only curvy smooth paths over path enabling smooth paths of small curvature. For example moving vertically in a hexagonal grid involves a great number of changes of direction in the grid based path. Despite this fact, the cost for curviness is still low for this kind of path.

We demonstrated the effectiveness of the proposed modification in different unstructured environments. The results show that, even though the grid cells are chosen small compared to the robots minimum curve radius, and also in presence of sharp obstacles, the resulting paths are smooth and satisfy the minimum curvature constraint. This property is independent

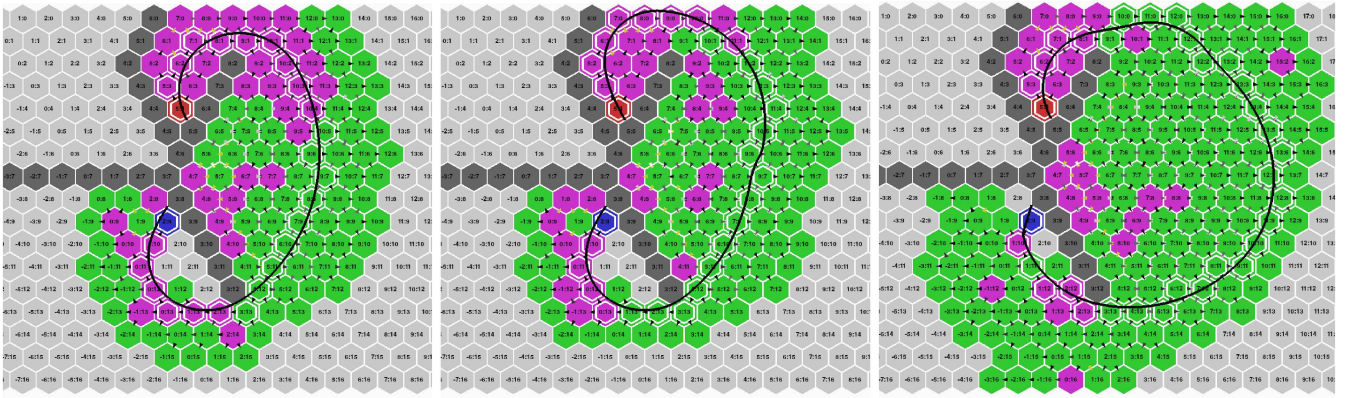

(a) Ribbon based cost (Sec. IV-A)  (b) Adapted ribbon based cost (Sec. IV-B)  (c) Manual cost distribution (Sec. IV-C)

Fig. 7: Resulting grid based paths and smooth paths with different curvature cost distributions.

of the orientation of the grid axes, as straight paths can be properly planned in any direction. By varying the weights on the curvature cost, the resulting path can be tuned to be optimal in length, in curvature or to include a minimum number of changes of direction.

The implementation of the curvature constraints involves an extended structure for the tree of solution path candidates. Admissibility of a cell is evaluated by back-propagating this path multiple cells. This does not imperatively increase the number of iterations required to solve the path planning problem, but the number of operations in each step increases. Usually, to compare the performance of variants of the $A^*$ algorithm, the number of iterations is used as a measure [14]. In the proposed modification, this comparison would not be expressive. For a decent comparison, an efficient implementation of the data structures and graph operations is necessary.